\documentclass[11pt,a4paper]{article}
\usepackage{authblk}
\usepackage[hyperref]{eacl2021}
\usepackage{times}
\usepackage{latexsym}
\usepackage{multirow}

\usepackage{graphicx} 
\graphicspath{ {images/} }

\usepackage{tabularx}

\usepackage{microtype}

\aclfinalcopy 

\title{Named Entity Recognition in the Legal Domain using a Pointer Generator Network}

\author{Stavroula Skylaki}
\author{Ali Oskooei}
\author{Omar Bari}
\author{Nadja Herger}
\author{Zac Kriegman}
\affil{Thomson Reuters Labs}

\date{}

\begin{document}
\maketitle
\begin{abstract}
Named Entity Recognition (NER) is the task of identifying and classifying named entities in unstructured text. In the legal domain, named entities of interest may include the case parties, judges, names of courts, case numbers, references to laws etc. We study the problem of legal NER with noisy text extracted from PDF files of filed court cases from US courts. The “gold standard” training data for NER systems provide annotation for each token of the text with the corresponding entity or non-entity label. We work with only partially complete training data, which differ from the gold standard NER data in that the exact location of the entities in the text is unknown and the entities may contain typos and/or OCR mistakes. To overcome the challenges of our noisy training data, e.g. text extraction errors and/or typos and unknown label indices, we formulate the NER task as a text-to-text sequence generation task and train a pointer generator network to generate the entities in the document rather than label them. We show that the pointer generator can be effective for NER in the absence of gold standard data and outperforms the common NER neural network architectures in long legal documents.
\end{abstract}

\section{Introduction}

Named Entity Recognition (NER) is the task of identifying the span and the class of a Named Entity (NE) in unstructured text. NEs typically include but are not limited to persons, companies, dates, and geographical locations \citep{Sang2003a}. 

Legal NER is a central task in language processing of legal documents, especially for extracting key information such as the name of the parties in a case, the court name or the case number, or references to laws or judgements, to name a few. The extracted NEs could be integrated in legal research workflows for functionalities such as search, document anonymization or case summarization  thereby enabling and expediting insights for legal professionals \citep{Zhong2020}.

NER is commonly formalized as a sequence labeling task: each token of the document is assigned a single label that indicates whether the token belongs to an entity from a predefined set of categories \citep{Li2018a}. To create a training dataset in such a format the annotator is required to manually label each token in a sentence with the respective category. In this format, both the NE and the location of the NE in the source text are known. This format of training data is what we refer to hereafter as “gold standard” data. Obtaining the required voluminous gold standard data to train such models is, therefore, a laborious and costly task.  

In this paper, we perform NER in filed lawsuits in US courts. Specifically, we aim to identify the party names in each case, i.e. the names of the plaintiffs and the defendants, in a large collection of publicly available cases from more than 200 courts in different US jurisdictions. The party names have been identified by legal annotators but their exact location in the text is unknown. In this respect, we do not have access to “gold standard” training data even though the target NEs are available. This feature of our dataset introduces a key difference of our task to most NER tasks.

One solution to this problem is to generate the “gold standard” training data by searching for the locations of the known NEs in the source text . By performing this additional transformation to our data, we would be able to train sequence labeling NER models. For the following reasons, this solution is nontrivial. First, as our source text is also extracted from scanned PDF files ("image-only" PDFs), it contains Optical Character Recognition (OCR) mistakes and/or typos which may not be present in the target NEs. Second, besides the potential OCR errors at the character level, the closely spaced, two-column page layouts that can be often found as headers in the filed cases, represent an additional challenge for the OCR, which tends to concatenate the text across columns (Figure 1). In such cases, the tokens that make up the NEs in the source text may be intertwined with other words and/or sentences. Third, variations of the names may be also present in the source text and in our human-generated labels, such as presence of first and/or middle names whole or as initials and, to a lesser extent, typos.   

\begin{figure*}[h]
\caption{The first page of a complaint found in our dataset along with the extracted source text and the target entities provided by legal annotators. The label set (target entities) consists only of the party names and not of their location in the extracted source.}
\centering
\includegraphics[width=1\textwidth]{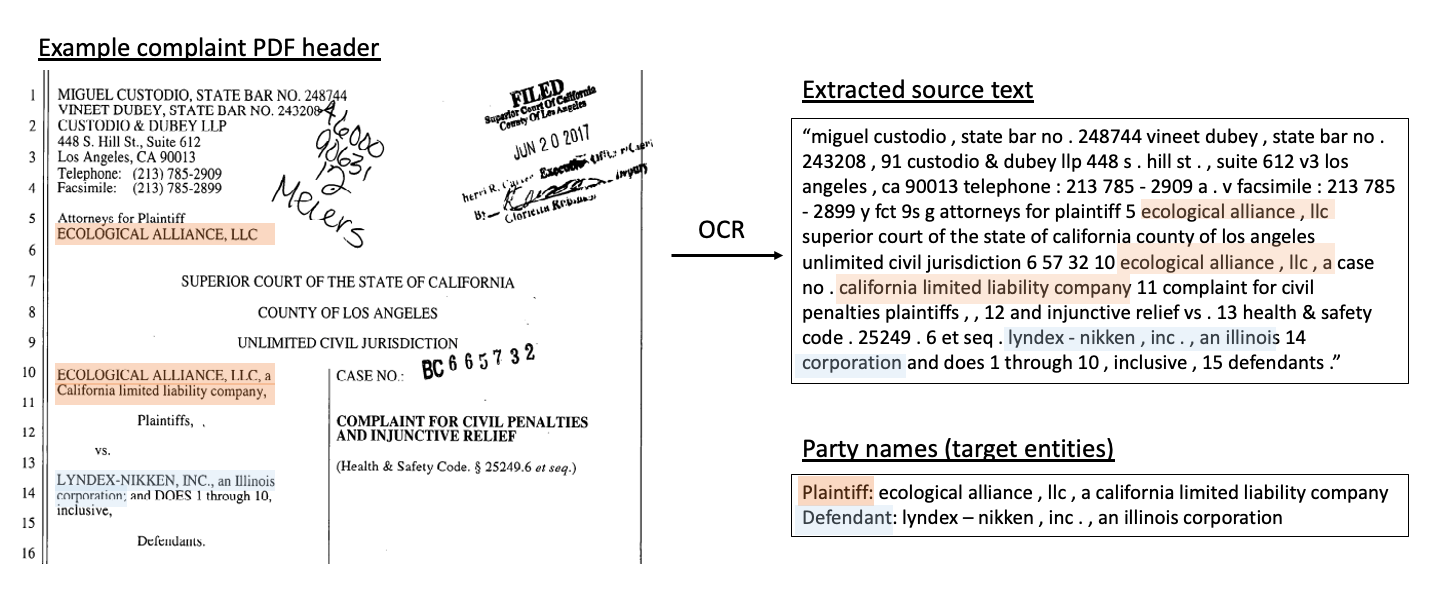}
\end{figure*}

To address some of the challenges imposed by the format of our training data and inspired by the work in the field of abstractive summarization, we propose to reformulate the NER task, not as a sequence labeling problem, but as a text-to-text sequence generation problem with the use of a pointer generator network \citep{Gu2016,See2017,Gehrmann2018}. With this reformulation, in contrast to sequence labeling, we do not require knowledge of the NE’s locations in the text as training labels. A recent study by \citet{Li2020} proposed a different formulation of the NER task as a question answering task and achieved state-of-the-art performance in a number of published NER datasets \citep{Li2020}. In this study, we adopt a hybrid extractive-abstractive architecture, based on recurrent neural networks coupled with global (i.e. the entire input document) attention and copying (or pointing) attention  mechanisms \citep{Gehrmann2018}. The proposed architecture can be successfully used for abstractive summarization since it can copy words from the source text via pointing and can deal effectively with out-of-vocabulary (OOV) words – words that have not been seen during training. Our approach is conceptually simple but empirically powerful and we show that the pointer generator outperforms the typical NER architectures in the case of noisy and lengthy inputs where the NE's location in the text is not known. 

In addition, we examine how our approach can be used for the related NER task of case number extraction. The case number is a unique combination of letters, numbers and special characters as a single token and are, therefore, particularly challenging for NER models as they are often dealt with as OOV words by the model. As in the party names task discussed above, in the case number task we do not have “gold standard” labels of the case number’s location in the text. We show that a character level sequence generation network can dramatically increase our ability to extract case numbers from the source text, compared to a word level sequence generation network.

The rest of the paper is organized as follows. In Section 2, we discuss related work in the field of NER in the legal domain. In Section 3, we describe our proposal of NER as a text-to-text sequence generation task in the absence of gold standard data and formulate the task in two ways: (i) as a combination of automatically labeling the NE's location and then using the conventional sequence labeling method for NER, and (ii) as a text-to-text sequence generation task where the NEs are directly generated as text. Section 4 presents our experimental design, results and analysis. Section 5 presents the case number case study. Finally, we conclude and discuss directions for future work.

\section{Related Work}

\subsection{NER in the legal domain}
There is a long history of research in the NER field ranging from statistical models such as Maximum Entropy Models \citep{Borthwick1998}, Hidden Markov Models \citep{DBLP:journals/ml/BikelSW99} or Conditional Random Fields (CRF) \citep{mccallum-li-2003-early}, to the current state-of-the-art deep neural network approaches based on bidirectional recurrent neural network (RNN) architectures often combined with a final CRF layer \citep{Ma2016b,Lample2016,DBLP:journals/corr/HuangXY15} as well as Transformers \citep{DBLP:conf/naacl/DevlinCLT19}. 

The vast majority of the developed NER approaches have been trained and evaluated on English texts from the general or news domains \citep{Li2018a}. This makes them less efficient for legal documents given the intricacies of the legal language, such as long complex sentences and domain specific vocabulary. For example, embeddings trained on general web crawling derived corpora might encounter many OOV words when used for the analysis of legal texts \citep{Rei2016}. 

Furthermore, most NER approaches operate at the sentence level which means that NER in long documents rely on effective sentence tokenizers. However, sentence boundary detection is challenging for legal text because of the variety of punctuation and syntax \citep{Sanchez2019} and the presence of idiosyncratic text formatting (e.g., the headers of legal documents where sentences may span multiple lines, contain tabular information and/or omit sentence markers) (Figure 1). This problem becomes especially exacerbated in the case of noisy, unstructured texts such as the ones produced by OCR systems where unavoidably mistakes create cascading effects to the downstream processing of the text \citep{DBLP:journals/ijdar/Lopresti09}. It is also worth noting that sentence level analysis prevents the model from benefiting from global information (a typical example is multiple mentions of the same named entities, e.g. the party names, occurring in different context in the document) \citep{Krishnan2006}. Figure 1 shows the first page of a representative complaint found in our dataset. The original PDF is displayed along with the extracted source text obtained with OCR. Here the name of the plaintiff is intertwined with the case number “case no” as a result of the OCR extraction of the double column format. Similarly, the words that make up the defendant name are interrupted by the token “14”. Finally, the plaintiff name appears in multiple locations in the text. 

\subsection{Attention-based sequence to sequence with pointing mechanism}
In the recent years, sequence-to-sequence neural networks have achieved great success in a number of natural language processing tasks, including but not limited to Machine Translation \citep{Bahdanau2014} and Text Summarization \citep{Gu2016,See2017,Gehrmann2018}. The backbone of these models is an encoder-decoder model \citep{Cho2014}, often coupled with an attention mechanism \citep{Bahdanau2014}. The attention mechanism enables the decoder to use a dynamically changing context, in the form of a context vector, in addition to the encoded state. Another mechanism, that has been particularly successful in the field of extractive summarization, is the pointing mechanism \citep{Gu2016,See2017,Gehrmann2018}. The pointing mechanism provides the model with a choice between generating tokens from the target vocabulary or copying where the decoder “copies” tokens directly from the source sequence. The pointing mechanism allows the model to be particularly effective when dealing with OOV words, as the decoder can copy words from the source sequence that are not in the vocabulary (e.g. very rare words). The pointer generator type network, and similar architectures such as CopyNet \citep{Gu2016}, combine the standard sequence to sequence model with a pointing ability. 

To the best of our knowledge, a pointer generator sequence to sequence model has not been applied to the NER task before. 

\section{NER as a text-to-text sequence generation task}
\subsection{The dataset}
Our dataset consists of 130,000 filed complaints from over 200 US courts spanning a time period of 9 years (2009-2018). A complaint is the initial formal action to initiate a lawsuit. The complaint is filed by the plaintiff and it contains information about the facts of the case, allegations against the defendant and the damages sought by the plaintiff. While complaints are made public by each US court, the collection of the data and the labeling of the target entities was performed in-house. The source text was extracted from the original PDF files using commercial OCR software (ABBYY FineReader). The page layout and text formatting of the complaints are highly variable as they are created by different law firms across many jurisdictions. The filed complaints are not following a template with predefined sections. An overview of the dataset can be found in Table 1.  

\begin{table*}
\centering
\begin{tabular} { l l l}
\hline \textbf{Characteristic} & \textbf{Statistic} & \textbf{Value} \\ \hline
Number of complaints & Total & 130000 \\
\hline
\multirow{2}{*} {Length of complaint (in tokens)} & 5th perc & 838 \\
& 50th perc & 2901 \\
& 95th perc & 16713 \\
\hline
\multirow{2}{*} {Number of party names in complaint} & Min & 2 \\
& Mean & 3 \\
& Max & 8 \\
\hline
Plaintiff names that can be located in the source text (perfect match) & Percent & 89.46 \\
\hline
Defendant names that can be located in the source text (perfect match) & Percent & 86.04 \\
\hline
\end{tabular}
\caption{\label{table1} Dataset overview for party names. }
\end{table*}

\subsection{Data format for NER as a sequence labeling task}
For each filed complaint in our dataset, annotators manually extracted a range of metadata including the party names. The exact indices of the extracted NEs in the source text are however unknown. We attempt to obtain the NE indices by aligning the identified NEs to the source text using the Knuth-Morris-Pratt string-matching algorithm to obtain a perfect match \citep{Knuth1977}. In Table 1, we show that the percentage of party names that can be aligned at least once to the source text is 89.46\% for plaintiff names and 86.04\% for defendant names). 

We use the identified indices of each NE to arrive at our gold dataset for the sequence labeling task. For this purpose we are using the BILOU format which identifies the Beginning, Inside and Last tokens of multi-token NEs as well as Unit-length NEs and Outside of NE words and has been shown to outperform other popular encoding options \citep{Ratinov2009}. So the NER sequence labeling task is formalized as following: given an input sequence of tokens $X={x_1,x_2,…,x_n }$, where $n$ denotes the length of the sequence, we need to find a sequence of labels $y={y_1,y_2,…,y_n }$, with $y \in Y$, where $Y$ is a predefined list of label types (e.g., B-plaintiff, B-defendant, I-plaintiff, etc). 

It is worth noting that multiple instances of the party names are often present in the source text and it is not possible to identify whether all instances have been correctly aligned, therefore a percentage of tokens is falsely labelled as not belonging to a NE (the equivalent of an ‘O’ label). 

In addition, we examine where in the source text the NEs can be found. In Figure 2, we show that the approx. 84\% of all aligned entities (both parties considered) can be found close to the start of the document. Therefore, we choose to use a cut-off of 1500 tokens (slope<=0.005), at which point the negligible gain in the percentage of aligned NEs is outweighed by the benefits of having a shorter input text for processing and training. Importantly, 1500 tokens sequence length is already much longer than the majority of public NER datasets on which most of the current SOTA results are reported (e.g. for CONLL2003 the average sentence length is approx. 14 tokens). It is worth considering that pre-trained models such as BERT \citep{DBLP:conf/naacl/DevlinCLT19}, which perform well at NER, have a limited input sequence length of 512 sub-tokens, and, therefore, the maximum sequence length is less than the suggested 1500 tokens above. 

\begin{figure}[h]
\caption{Percentage of NEs that can be aligned to the source text by perfect match (Knuth-Morris-Pratt) as a function of the source text length (in tokens). The vertical dash line indicates the selected length threshold (1500 tokens, point where $slope \leq 0.005$)}
\centering
\includegraphics[width=0.5\textwidth]{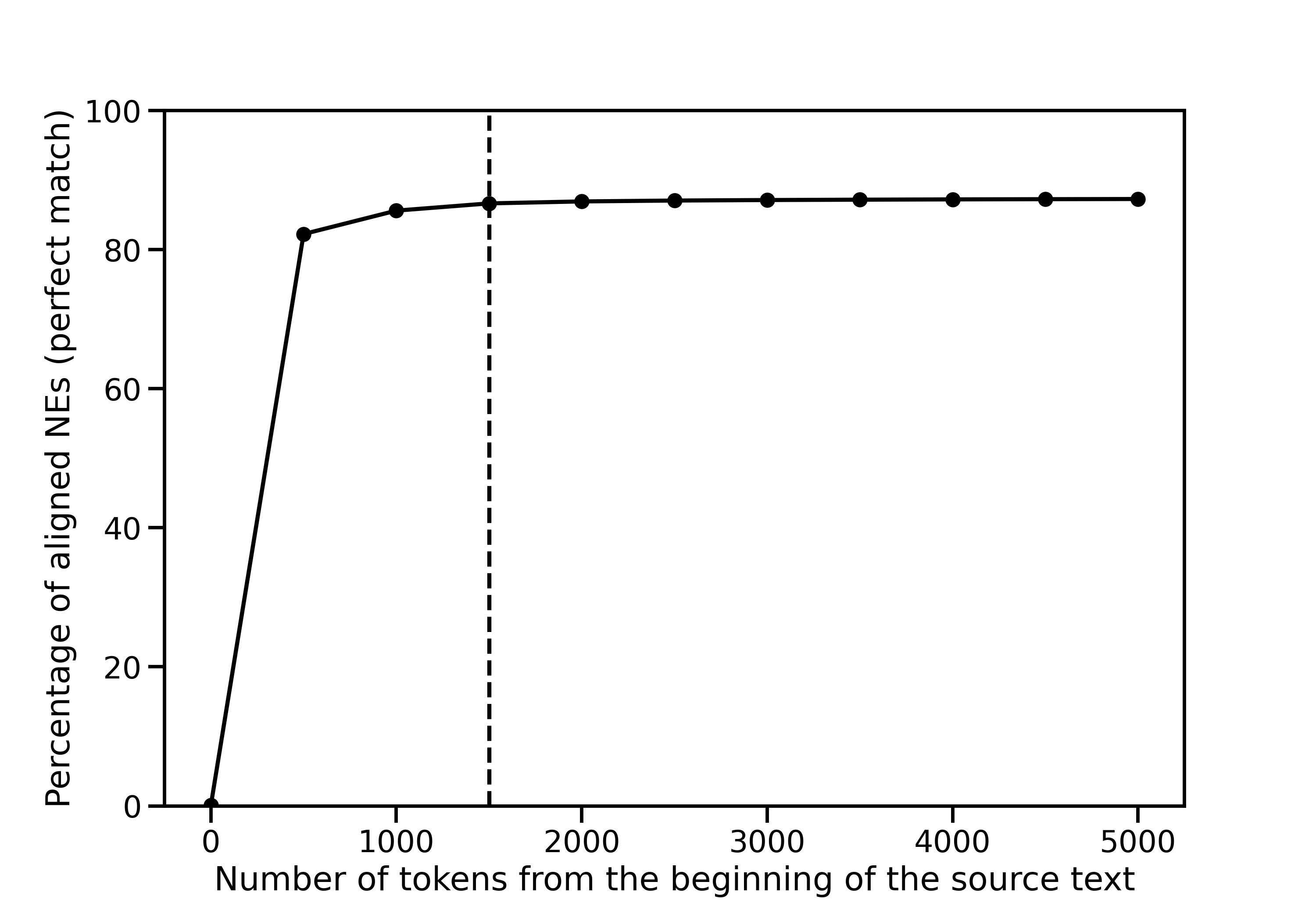}
\end{figure}

\subsection{Data format for NER as a text-to-text sequence generation task}
We now formalize the NER task as a sequence generation task as follows: given an input sequence of tokens $X=\{x_1,x_2,…,x_n \}$, where $n$ denotes the length of the sequence, we need to find every entity in $X$, and assign a label $y \in Y$ to it, where $Y$ is a predefined list of label types (in our case study the two type of party names i.e. plaintiff or defendant). We formulate our target sequence as a series of labels followed by the entity tokens and, for each entity in $X$, we obtain: $<y> x_{start},x_{start+1},…,x_{end-1},x_{end}$
e.g. $<$plaintiff$>$ jane doe $<$defendant$>$ acme corporation

\section{Experiments}
\subsection{Pointer generator model}
We have used the hybrid extractive-abstractive pointer generator type of network previously described by See et al. \citep{See2017} as implemented in the PyTorch version of the OpenNMT sequence to sequence toolkit \citep{Gehrmann2018,Klein2018}. This is essentially a bidirectional long short-term memory (LSTM) encoder-decoder architecture with Bahdanau attention \citep{Bahdanau2014} and pointing mechanism \citep{Gu2016}. We used two hidden layers of 512 size and word embedding dimension of 100. During inference we used beam search size 10 to which we applied length normalization according to Wu et al. (alpha=0.9) \citep{Wu2016}. We used the summary specific coverage penalty described in \citet{Gehrmann2018} with beta = 5. This penalty blocks sequence generation with repetitions. Additionally, we restricted the beam search to never repeat bigrams.

\subsection{Baseline NER models}
We compared the performance of our proposed approach with the following commonly used neural network architectures for NER: 
\begin{itemize}
\item spaCy, an open-source, NLP library for a variety of tasks including NER which uses subword “Bloom” embeddings and deep convolutional neural networks (CNN) with residual connections \citep{Honnibal2017}. 
\item BiLSTM-CRF as described in Lample et al. \citep{Lample2016}. 
\item DistilBERT-Tagger, similar to Devlin et al. \citep{DBLP:conf/naacl/DevlinCLT19} but using the pre-trained DistilBERT, a distilled version of BERT with increased speed and reduced size while retaining 97\% of BERT performance \citep{Sanh2019}. 
\end{itemize}

\subsection{Results}

\begin{table*}
\centering
\begin{tabular} { l l l l l l}
\hline \textbf{Model} & \textbf{Finetuning} & \textbf{Type}  & \textbf{Precision} & \textbf{Recall} & \textbf{F1-Score} \\ \hline
spaCy & No & CNN-SeqLabel & 8 & 49.4 & 13.8 \\
BLSTM-CRF & No & RNN-SeqLabel & 60.9 & 76.7 & 67.9 \\
DistilBERT & Yes & Transformer-SeqLabel & 75.3 & 70.8 & 73 \\
\hline
\textbf{Pointer Generator} & \textbf{No} & \textbf{RNN-SeqGen} & \textbf{77.3} & \textbf{71.8} & \textbf{74.5}\\
\hline
\end{tabular}
\caption{\label{table2} Results from the tested architectures. The first section shows the performance of NER architectures i.e. sequence labeling models. The second section show the performance of the proposed Pointer Generator model, a text-to-text sequence generation approach. }
\end{table*}

Table 2 presents the comparison between the proposed sequence to sequence model and the baseline NER models. For spaCy, our motivation was to explore the performance of a popular out-of-the-box NER model in our NER task which is specialized in the legal domain. We used the large English model, trained on written web text (en\_core\_web\_lg). We retrieved all entities labeled as person “PER” or organization “ORG” which could be the case parties. Even though, this approach cannot clearly label the plaintiff and defendants so that subsequent classification of the retrieved entities is required, it can however give us an indication of the performance of the model by looking at the recall of the model i.e. high recall means that the party names are indeed in the list of retrieved entities. Table 2 shows that the pre-trained spaCy model cannot compete in this specialized legal NER task and further training is required. In addition, the Pointer Generator clearly outperforms the BLSTM-CRF benchmark (+6.6\%), and the finetuned DistilBERT tagging model (+1.5\%) in terms of F1-Score. 
One reason for the better performance of the Pointer Generator compared to the traditional NER architectures may lie on its ability to extract entities that have been fragmented because of document formatting. For example, in the complaint shown in Figure 1, the model correctly extracts the name of the plaintiff and the name of the defendant even though both names are intertwined with unrelated tokens e.g. “case no” and “14”. In addition, the model relies on the pointing mechanism in 36\% of the cases which indicates how often the pointer generator is able to deal with OOV words by pointing to the source text.  

We have evaluated the models using precision, recall and F1-score:

\[ Precision = |\{Rel\} \cap \{Retr\}| / |\{Retr\}| \]
\[ Recall = |\{Rel\} \cap \{Retr\}| / |\{Rel\}| \]
\[ F1 = 2 * (Pr * Re) / (Pr + Re) \]
where Rel is the number of relevant names i.e. names that belong to the party entities and Retr is the number of retrieved names. Pr stands for Precision and Re stands for Recall. Relevant and retrieved names in the above formulas are compared using perfect match (excluding punctuation). 

\subsection{Ablation Studies}
Next, we investigate whether the observed improvement of the Pointer Generator is the result of the longer input allowed in the specific model compared to DistilBERT or the reformulation of the NER task as a sequence generation task. For this purpose, we compared the performance of the models in the subset of complaints where all entities can be mapped according to the typical NER tagging scheme and where the length of the source text is smaller than 512 tokens, which approximates the input limits of DistilBERT.  In Table 3, we show that the finetuned DistilBERT-Tagger outperforms the Pointer Generator in the case of short text inputs (+6.9\% F1 Score). This indicates that the major advantage of the Pointer Generator is its capacity to deal with significantly longer text inputs and to overcome formatting issues.

\begin{table*}
\centering
\begin{tabular} { l l l l l l}
\hline \textbf{Model} & \textbf{Finetuning} & \textbf{Type}  & \textbf{Precision} & \textbf{Recall} & \textbf{F1-Score} \\ \hline
\textbf{DistilBERT} & \textbf{Yes} & \textbf{Transformer-SeqLabel} & \textbf{88.8} & \textbf{89} & \textbf{88.9} \\
\hline
Pointer Generator & No & RNN-SeqGen & 77.3 & 71.8 & 74.5\\
\hline
\end{tabular}
\caption{\label{table3} Results on the effect of input text length to the performance of the models. Here the model performance is calculated on the subset of the testset where the input text length < 512 tokens and the entities can be perfectly mapped to the source text in accordance to the typical NER labeling scheme.}
\end{table*}

In addition, we investigate whether the improvement we see in the performance of the Pointer Generator is due to the reformulation of the NER task as text-to-text sequence generation task. We therefore tested our approach against BART-CNN, a large-scale pretrained language model fine-tuned on the summarization task using the CNN/Daily Mail dataset \citep{Lewis2020}. We further finetuned the BART-CNN model with our dataset. Table 4 presents the comparison between BART-CNN and the proposed Pointer Generator. The Pointer Generator outperforms BART-CNN by a margin of +3.8\% F1 score which may indicate that BART-CNN which has been pre-trained in largely dissimilar dataset of news articles did not manage to adapt to the NER task in the legal domain. 

\begin{table*}
\centering
\begin{tabular} { l l l l l l}
\hline \textbf{Model} & \textbf{Finetuning} & \textbf{Type}  & \textbf{Precision} & \textbf{Recall} & \textbf{F1-Score} \\ \hline
BART-CNN & Yes & Transformer-SeqGen & 75.6 & 66.4 & 70.7 \\
\hline
\textbf{Pointer Generator} & \textbf{No} & \textbf{RNN-SeqGen} & \textbf{77.3} & \textbf{71.8} & \textbf{74.5}\\
\hline
\end{tabular}
\caption{\label{table4} Results on different models where the NER task is reformulated as a text-to-text sequence generation task.}
\end{table*}

\section{Case number case study}
Here we present and discuss a particular challenging problem for traditional NER methods. Case numbers are identification codes assigned to each complaint. Depending on the court, the case number can be of varying length and format and may include letters, numbers, a combination of letters and numbers or even special characters such as colon (:), slash (/) or a dash (-) (e.g. 1:16cv00678, 0002088/2016, etc.). As a result of this formatting diversity, each case number is in essence unique. The uniqueness of case numbers is problematic for conventional vocabulary-based neural NER models where a finite vocabulary of most frequent words (i.e. tokens) is built from the data and used to develop a model that is able to generalize on unseen data containing the same vocabulary \citep{Luong2014}. It is guaranteed that any new unseen legal case will have a case number that is OOV and will be represented by the ‘unk’ umbrella token for OOV words. This limitation hinders the ability of any conventional neural NER model to generalize well to unseen data.

To address the OOV problem, we adopted a two-pronged approach: 1) using a sequence to sequence model with pointing ability such as the pointer generator network \citep{See2017}. 2) using the same model with character tokens instead of word tokens. Pointer Generator networks have been shown to be successful at tackling the OOV problem by determining and sampling from a probability distribution over an “extended vocabulary” that includes the usual most frequent vocabulary in addition to all words in the input document. The addition of the input words to the vocabulary for each document results in the model having access to all words in the document even if they are rare words and thereby tackling the OOV problem. Figure 3 shows the prediction accuracy versus Levenshtein distance between the predicted and true case numbers for the pointer generator model with word tokens. As shown in the bar plot the model performs poorly with an accuracy of 49\% despite the model having access to case numbers in the input document. We postulate that this is caused by the generator part of the pointer generator network dominating the pointer part and preventing the model from pointing to the rare input tokens in favor of generating a frequently occurring word from the vocabulary. 

\begin{figure}[h]
\caption{Accuracy of the pointer generator sequence to sequence model on extracting case numbers when using word tokens and character tokens. Accuracy is calculated for various Levenshtein error tolerances. As shown in the dark grey bar plots, using character tokens results in a model that makes fewer than 1 character errors in over 80\% of cases while that ratio is 50\% for the word token model.}
\centering
\includegraphics[width=0.5\textwidth]{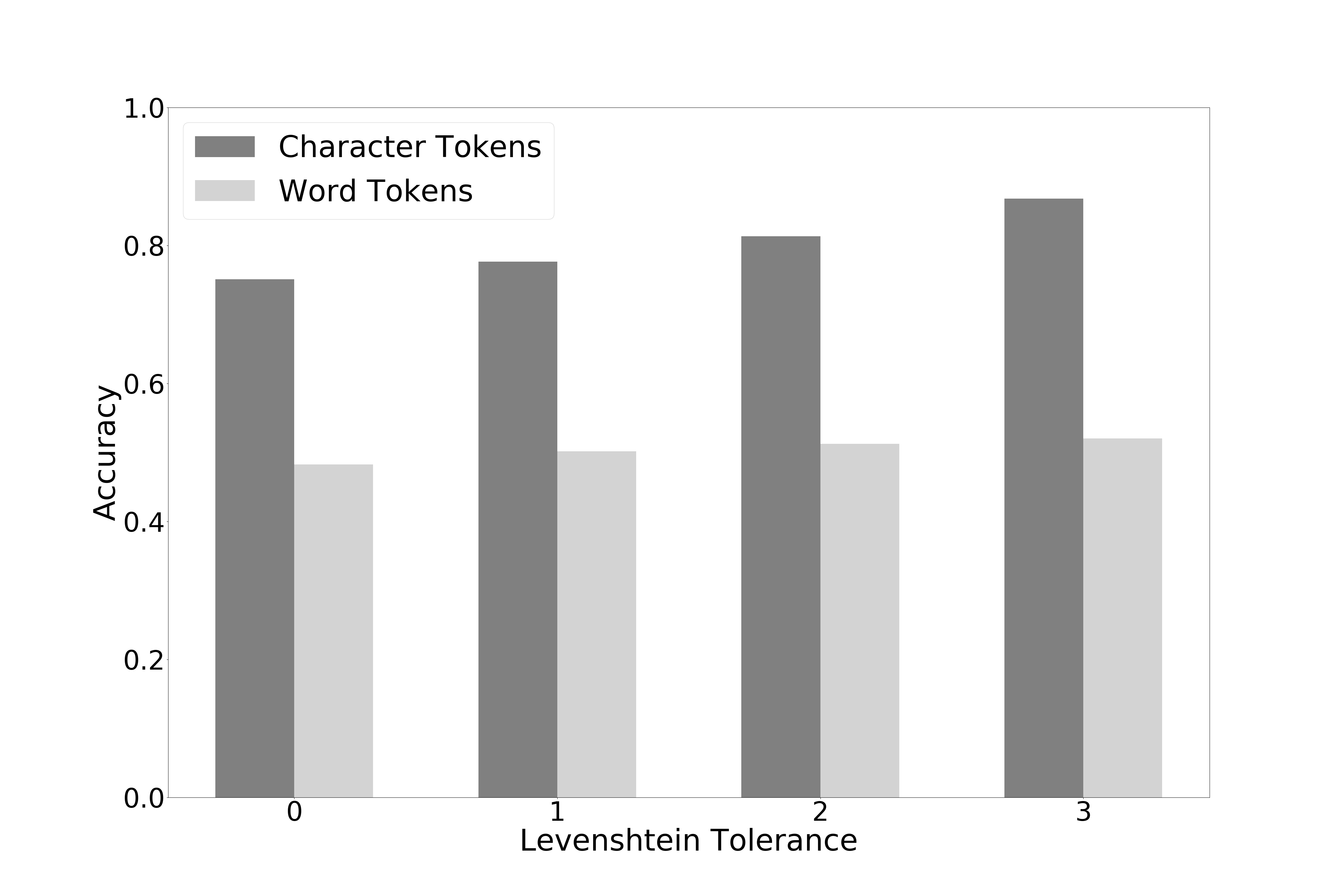}
\end{figure}

Alternatively, we trained a pointer generator with character tokens. The vocabulary, in this case, is reduced to a finite number of characters (alphabet plus the special characters) and as such pointing is meaningless and the model is reduced to a standard sequence to sequence model with attention. In this scenario, the input case text is tokenized into characters and fed into the model which learns to generate the case number character by character. The character tokens eliminate the OOV problem by enabling the generation of any newly encountered terms character by character. Figure 3 also shows the accuracy versus Levenshtein distance of predicted and true case numbers using the character-based model. We achieve a 50\% improvement in model accuracy using character tokens. Our results demonstrate suitability of character-based sequence to sequence models for NER tasks on rare and OOV words such as the legal case numbers and various other types of unique identifiers for a large number of samples.

\section{Conclusions}

This work presents a simple yet powerful reformulation of the NER task as a text-to-text sequence generation task by applying a pointer generator network, a model architecture that have been predominantly used in the NLP field of summarization. There are several key advantages in the proposed formalization:  (1) there is no need to acquire “gold” data for the NER task when only the target NEs are known but not their indices in the source text, (2) the Pointer Generator network outperforms popular sequence-labeling architectures at the NER task in the case of longer text inputs, and (3) the Pointer Generator is able to accurately generate NEs that are corrupted due to OCR errors in extracting the two-column formatted text. In the future, we would like to explore the capacity of the Pointer Generator to extract additional types of NEs.

\bibliography{legalNERwtPNG}
\bibliographystyle{acl_natbib}

\end{document}